\documentclass{article}

\PassOptionsToPackage{numbers, compress}{natbib}

\usepackage[preprint]{neurips_2024}

\usepackage{algorithm}
\usepackage{algpseudocode}
\usepackage[utf8]{inputenc}
\usepackage[T1]{fontenc}
\usepackage{hyperref}
\usepackage{url}
\usepackage{longtable}
\usepackage{siunitx}
\usepackage{booktabs}
\usepackage{amsfonts}
\usepackage{nicefrac}
\usepackage{microtype}
\usepackage{xcolor}
\usepackage{amsmath}
\usepackage{graphicx}
\usepackage{geometry}
\usepackage{authblk}
\usepackage[most]{tcolorbox}
\usepackage{enumitem}
\usepackage{xcolor}
\title{Scientific Discovery under Imperfect Evaluators: Diversity-Aware Neuro-Symbolic Search for Solvent Design}

\author[1,2,*]{Jiangyu Chen}
\author[2,*]{Huawei Zhou}
\author[1]{Zhou Zhang}
\author[1,$\dagger$]{Tianfan Fu}
\author[2,$\dagger$]{Ruzhi Zhang}

\affil[1]{State Key Laboratory for Novel Software Technology at Nanjing University, School of Computer Science, Nanjing University}
\affil[2]{Suzhou Laboratory}

\affil[*]{Equal contribution.}
\affil[$\dagger$]{Corresponding author.}

\begin{document}

\maketitle

\begin{abstract}
Scientific discovery in chemical formulation often unfolds under imperfect evaluators: wet-lab validation is expensive, combinatorial design spaces are large, and available scores only partially reflect downstream experimental performance. In such settings, optimizing a single proxy score can narrow exploration too early and miss experimentally valuable candidates. We study this challenge in solvent design, a mixed discrete--continuous problem that requires selecting component sets and optimizing mixing ratios under explicit physicochemical constraints.

We propose \textbf{AI4S-SDS} (\textbf{AI for Science--Solvent Design System}), a diversity-aware neuro-symbolic framework for discovery under imperfect evaluators. LLMs act as chemistry-informed hypothesis generators over discrete formulation topologies, while a differentiable physics-informed optimization module refines mixing ratios and enforces feasibility. To reduce search collapse toward evaluator-preferred patterns, the framework combines sibling-aware local diversification with memory-driven global planning.

Experiments show that \textbf{AI4S-SDS} maintains 100\% compliance with the explicit physicochemical constraints adopted in our framework and substantially improves exploration diversity over score-centric baselines. In preliminary lithography tests, representative formulations identified by the framework exhibit better qualitative pattern definition than a commercial baseline under the tested conditions, even when they are not top-ranked by the scoring function. Taken together, the results suggest that, in discovery settings with imperfect evaluators, diversity-aware search can be more effective than proxy-score maximization alone for surfacing experimentally promising candidates.
\end{abstract}

\section{Introduction}
Many scientific search problems are conducted under imperfect evaluators rather than reliable gold metrics. In solvent design, wet-lab validation is costly, candidate spaces are combinatorial, and available scoring functions capture only part of the behavior that matters in practice. Even in our setting, restricting the search to 50 candidate solvents and mixtures of 2 to 5 components already yields millions of discrete combinations before continuous mixing ratios are considered. At the same time, formulation design is inherently mixed discrete--continuous: one must decide which components to include and in what proportions, while satisfying physicochemical, safety, and engineering constraints. This makes solvent discovery a discovery-oriented search problem in which the objective is not simply to maximize a scalar proxy score, but to surface constraint-satisfying, diverse, and experimentally promising candidates for downstream validation.

In such settings, large language models (LLMs) are useful not as exact numerical optimizers, but as semantic generators over structured chemical design spaces. They can analyze property patterns, reason over component roles, and propose plausible formulation topologies that go beyond unguided trial-and-error. In our framework, this role is explicit. LLMs generate discrete formulation hypotheses, while continuous ratios are refined by a differentiable physics-informed optimization module. This separation matches the chemistry workflow encoded in our prompts: qualitative reasoning guides candidate construction, whereas numerical feasibility is handled by a dedicated optimization module grounded in HSP geometry, separation dimensions, and component functional roles.

However, most existing LLM-based agentic systems are not designed for discovery under imperfect evaluators. Prior work has shown that LLMs can coordinate tools, follow procedures, and support symbolic planning, but formulation discovery requires something more specific: long-horizon exploration over large combinatorial spaces together with continuous, feasibility-constrained optimization. More importantly, when the evaluator itself is incomplete, aggressively optimizing a single scalar score can cause the search to collapse onto evaluator-preferred templates and overlook chemically distinct but experimentally promising solutions.

This perspective motivates our central design choice: diversity should be treated as a structural property of the search process rather than as a cosmetic add-on. Accordingly, we propose \textbf{AI4S-SDS} (\textbf{AI for Science--Solvent Design System}), a diversity-aware neuro-symbolic framework that combines chemistry-guided hypothesis generation, differentiable physicochemical feasibility checks, sibling-aware local diversification, and memory-driven global planning. Empirically, this design maintains full compliance with the explicit physicochemical constraints adopted in this work, increases exploration entropy from 3.53 to 4.37 relative to Naive MCTS, and identifies formulations absent from baseline search trajectories under the same budget. Preliminary lithography experiments further show that representative candidates from these diverse regions can exhibit more favorable qualitative pattern definition than a commercial baseline under the tested conditions, even when they are not ranked highest by the scoring function.

Our main contributions are summarized as follows:
\begin{itemize}
    \item \textbf{A discovery-oriented perspective for solvent design under imperfect evaluators:} We formulate solvent design as a mixed discrete--continuous scientific search problem in which reliable gold metrics are unavailable, wet-lab validation is expensive, and scalar evaluators serve only as imperfect proxies of downstream experimental performance. This perspective shifts the objective from pure score maximization toward the discovery of constraint-satisfying, diverse, and experimentally promising candidate sets.
    \item \textbf{A diversity-aware neuro-symbolic framework for chemistry-guided search:} We propose \textbf{AI4S-SDS} (\textbf{AI for Science--Solvent Design System}), a framework that decouples discrete formulation generation from continuous ratio optimization. LLMs act as chemistry-informed hypothesis generators over solvent topologies, while a differentiable physics-informed optimization module refines proportions and enforces feasibility. The framework further combines sibling-aware local diversification with memory-driven global planning to mitigate search collapse and promote exploration across multiple feasible chemical regions.
    \item \textbf{Empirical evidence that diversity-aware search can outperform score-centric search in discovery settings:} We show that AI4S-SDS maintains 100\% compliance with the explicit physicochemical constraints adopted in this work while substantially improving exploration diversity relative to baseline agents. Although diversity-aware search trades off some short-term proxy score, it uncovers formulations in previously unexplored chemical subspaces, and preliminary lithography results indicate that such candidates can still exhibit favorable qualitative experimental behavior.
\end{itemize}
\section{Related Work}

Recent progress at the intersection of large language models (LLMs), autonomous agents, and scientific computing has advanced computational chemistry and materials discovery.
This section reviews three converging research directions most relevant to our work:
(1) LLM-based agentic systems for scientific discovery,
(2) structured reasoning and search architectures, and
(3) optimization methods for chemical design.
Across these lines of work, the central gap is not merely the absence of stronger generators, but the lack of frameworks that combine chemistry-informed proposal generation, physics-constrained feasibility, and diversity-preserving search in discovery settings where proxy scores are incomplete.

\subsection{LLM-based Agents for Scientific Discovery}

The emergence of agentic LLM architectures marks a shift from passive text generation to autonomous scientific reasoning and action. Systems such as \textsc{Coscientist}~\cite{boiko2024autonomous} demonstrate that LLMs can coordinate planning, tool use, and iterative refinement in closed-loop scientific workflows, while tool-augmented systems such as \textsc{ChemCrow}~\cite{bran2024chemcrow} reduce hallucinations and improve chemical validity through external calculators, databases, and safety checks. Domain-specific chemistry tuning, exemplified by \textsc{ChemLLM}~\cite{zhang2024chemllm}, further shows that chemical reasoning can be strengthened through specialized instruction data.

\noindent\textbf{Limitation.}
Despite their success, existing agentic systems mainly operate in discrete, procedural, or symbolic spaces. They show that LLMs can execute workflows and interact with tools, but they do not address a key challenge of formulation discovery: when no reliable gold metric is available, the LLM is most useful not as a direct optimizer, but as a chemistry-informed hypothesis generator embedded in a search loop that must also preserve diversity and enforce physicochemical feasibility. Existing systems therefore fall short of supporting discovery-oriented search over mixed discrete–continuous formulation spaces.

\subsection{Structured Reasoning and Search with LLMs}

To overcome the limitations of linear Chain-of-Thought (CoT) prompting, recent work has introduced structured reasoning architectures that impose explicit search over the LLM’s generative space.
Tree of Thoughts (ToT) and Graph of Thoughts (GoT)~\cite{besta2024graph,yao2023tree} generalize CoT into tree or graph topologies, enabling branching, backtracking, and aggregation of intermediate reasoning states.

More recently, Reasoning via Planning (RAP) integrates Monte Carlo Tree Search (MCTS) with LLMs~\cite{lu2023rap}.

It is worth noting that MCTS has been successfully applied to chemical synthesis planning even prior to the LLM era, as demonstrated by Segler et al.~\cite{segler2018planning}.
By performing rollouts and balancing exploration and exploitation, these search-based methods improve long-horizon planning performance in complex reasoning benchmarks.

\noindent\textbf{Limitation.}
While general-purpose mechanisms like MemGPT~\cite{packer2023memgpt} or Reflection~\cite{shinn2023reflexion} address context limits via memory retrieval, they operate primarily on textual semantics.
They lack the structured, state-based abstraction required to maintain the logical integrity of a scientific search tree over infinite horizons.

\subsection{Optimization in Chemical Design}

Chemical formulation design fundamentally requires optimization over mixed discrete–continuous spaces.
Bayesian optimization is a standard choice for expensive black-box objectives, but its performance degrades in high-dimensional settings with combinatorial explosion.~\cite{frazier2018tutorial}.
However, its performance degrades in high-dimensional settings, where kernel-based surrogates struggle with combinatorial explosion.

Prior to LLMs, deep generative models such as Variational Autoencoders (VAEs)~\cite{gomez2018automatic} were widely adopted to map discrete molecular graphs to continuous latent spaces for optimization.
While effective for generation, these continuous embedding methods often lack the explicit semantic reasoning required for complex, multi-constraint formulation tasks.

Genetic Algorithms (GA) are commonly applied to discrete molecular spaces~\cite{sheridan2011ga}, but they are notoriously sample-inefficient.

Beyond evolutionary strategies, Deep Reinforcement Learning (DRL) has emerged as a powerful paradigm for de novo design~\cite{olivecrona2017molecular, popova2018deep}.
Methods like REINVENT optimize molecular structures by treating generation as a policy learning problem.
However, DRL approaches often require extensive pre-training and struggle with sample efficiency in low-data regimes typical of experimental formulation discovery.

Recently, large language models have also been proposed as direct optimizers (OPRO) by iteratively prompting for better solutions~\cite{yang2024large}.
However, these methods operate purely on textual prompts and lack the numerical precision provided by gradient-based physical simulation.
Finally, the growing field of Physics-Informed Machine Learning (PIML) seeks to integrate physical laws into deep learning frameworks~\cite{karniadakis2021physics}.
However, most PIML approaches focus on solving differential equations rather than guiding discrete combinatorial search in agentic systems.

\noindent\textbf{Limitation and positioning.}
Existing optimization paradigms each solve only part of the problem. In formulation discovery, the central difficulty is often upstream of optimization itself: the scalar evaluator is only a proxy for real experimental value. This makes pure score-driven optimization brittle. Our work is positioned differently. Rather than assuming that better optimization of a scalar reward is sufficient, we treat solvent discovery as a discovery-oriented search problem under imperfect evaluators, and therefore combine chemistry-guided hypothesis generation, differentiable physicochemical feasibility checks, and diversity-aware search dynamics in a single iterative computational framework. Table~\ref{tab:method_comparison} summarizes this distinction. Compared with prior methods, \textbf{AI4S-SDS} is not only an agentic search system with physics grounding; more importantly, it is designed for discovery under imperfect evaluators, where broad exploration across multiple feasible candidate families is often more valuable than maximizing a single proxy score.

\begin{table}[h]
\centering
\caption{Comparison of related methods and \textbf{AI4S-SDS}.}
\label{tab:method_comparison}
\resizebox{0.95\columnwidth}{!}{
\begin{tabular}{lcccccc}
\toprule
\textbf{Method} &
\textbf{Agentic} &
\textbf{Search / Planning} &
\textbf{Sparse Memory} &
\textbf{Diversity Control} &
\textbf{Physics-based Opt.} &
\textbf{Iterative Loop} \\
\midrule
ChemLLM \cite{zhang2024chemllm} & \checkmark & -- & -- & -- & -- & -- \\
ChemCrow \cite{bran2024chemcrow} & \checkmark & -- & -- & -- & \checkmark (Tools) & -- \\
Coscientist \cite{boiko2024autonomous} & \checkmark & Limited & -- & -- & -- & \checkmark \\
ToT / GoT \cite{yao2023tree} & \checkmark & \checkmark & -- & -- & -- & -- \\
RAP \cite{lu2023rap} & \checkmark & \checkmark (MCTS) & -- & -- & -- & -- \\
\midrule
\textbf{AI4S-SDS} (Ours) &
\checkmark &
\checkmark (Sparse MCTS) &
\checkmark &
\checkmark &
\checkmark (Diff. Physics) &
\checkmark \\
\bottomrule
\end{tabular}
}
\end{table}

\section{Problem Formulation}

We formulate the solvent design task as a hierarchical optimization problem
that jointly considers discrete compositional structure and continuous mixing ratios.
The objective is to identify solvent formulations that achieve high solubility
with a target photoresist while satisfying physicochemical and safety constraints.

\subsection{The Chemical Search Space}

Let $\mathcal{S} = \{s_1, s_2, \dots, s_N\}$ denote a candidate pool of $N = 50$ commercially available solvents.
A solvent formulation is formally defined as a tuple $(\mathcal{M}, \boldsymbol{\phi})$, where:

\begin{itemize}
    \item \textbf{Discrete Topology:}
    $\mathcal{M} \subset \mathcal{S}$ represents the subset of selected components, subject to the sparsity constraint $2 \le |\mathcal{M}| \le 5$.
    
    \item \textbf{Continuous Geometry:}
    $\boldsymbol{\phi} \in \mathbb{R}^{|\mathcal{M}|}$ denotes the \textbf{volume fractions} associated with components in $\mathcal{M}$, satisfying the simplex constraint $\sum_{i \in \mathcal{M}} \phi_i = 1$ and $\phi_i > 0$.
\end{itemize}

The dimensionality of the continuous variable $\boldsymbol{\phi}$ is conditioned on the selected topology $\mathcal{M}$, inducing a variable-dimensional, non-convex, and mixed discrete--continuous search space.
This hierarchical structure poses a significant challenge for traditional gradient-only optimization methods, which typically assume a fixed parameter space.

\subsection{Theoretical Basis: Hansen Solubility Parameters}

To quantify the physicochemical affinity between a solvent mixture ($\text{mix}$) and a target material ($\text{target}$), we adopt the Hansen Solubility Parameters (HSP) theory.
Each material is characterized by a triplet vector $\boldsymbol{\delta} = (\delta_d, \delta_p, \delta_h)$, corresponding to atomic dispersion, molecular dipolar, and hydrogen-bonding interactions, respectively.

For a mixture defined by volume fractions $\boldsymbol{\phi}$, the effective HSP parameters are derived via the linear mixing rule~\cite{hansen2007hansen}:
\begin{equation}
    \boldsymbol{\delta}_{\text{mix}}(\boldsymbol{\phi}) = \sum_{i \in \mathcal{M}} \phi_i \boldsymbol{\delta}_{i}.
\end{equation}
The affinity metric, commonly referred to as the HSP distance ($R_a$), is defined as:
\begin{equation}
R_a(\text{mix}, \text{target}) =
\sqrt{
4(\delta_d^{\text{mix}} - \delta_d^{\text{target}})^2
+ (\delta_p^{\text{mix}} - \delta_p^{\text{target}})^2
+ (\delta_h^{\text{mix}} - \delta_h^{\text{target}})^2
}.
\end{equation}
Note that the factor of $4$ in the dispersion term is empirical, accounting for the steric nature of dispersion forces.
A smaller $R_a$ indicates higher solubility.
Crucially, Eq.~(1) and (2) establish a differentiable mapping from the formulation ratio $\boldsymbol{\phi}$ to the property space, enabling end-to-end gradient-based guidance.

\subsection{Optimization Objective}

Our goal is to identify an optimal formulation $(\mathcal{M}^*, \boldsymbol{\phi}^*)$ that satisfies three competing criteria:
\begin{enumerate}
    \item \textbf{Solubility Efficiency:} Maximize affinity to the target photoresist ($R_a(\text{mix}, \text{target}) \rightarrow 0$);
    \item \textbf{Selectivity \& Safety:} Ensure the formulation remains inert to the protective layer ($R_a(\text{mix}, \text{protect}) > R_{\text{safe}}$);
    \item \textbf{Physical Validity (PV):} Compliance with the explicit physicochemical and thermodynamic constraints adopted in this work, including boiling-point hierarchy and flash-point safety.
\end{enumerate}

Instead of computationally expensive Pareto front enumeration, we formulate a unified scalar objective via a hybrid penalty method:
\begin{equation}
\min_{\mathcal{M}, \boldsymbol{\phi}} \;
\mathcal{J} =
L_{\text{hybrid}}(\mathcal{M}, \boldsymbol{\phi})
+ \lambda \cdot \mathcal{C}(\mathcal{M}, \boldsymbol{\phi}),
\end{equation}
where $L_{\text{hybrid}}$ represents a normalized loss balancing relative selectivity and absolute solubility, and $\mathcal{C}(\cdot)$ denotes a differentiable penalty term encoding physical constraints.
The explicit derivation of $L_{\text{hybrid}}$ is detailed in Section~4.

\section{Methodology}

\subsection{Overview and Design Principle}
\begin{figure}[ht]
    \centering
    \includegraphics[width=\columnwidth]{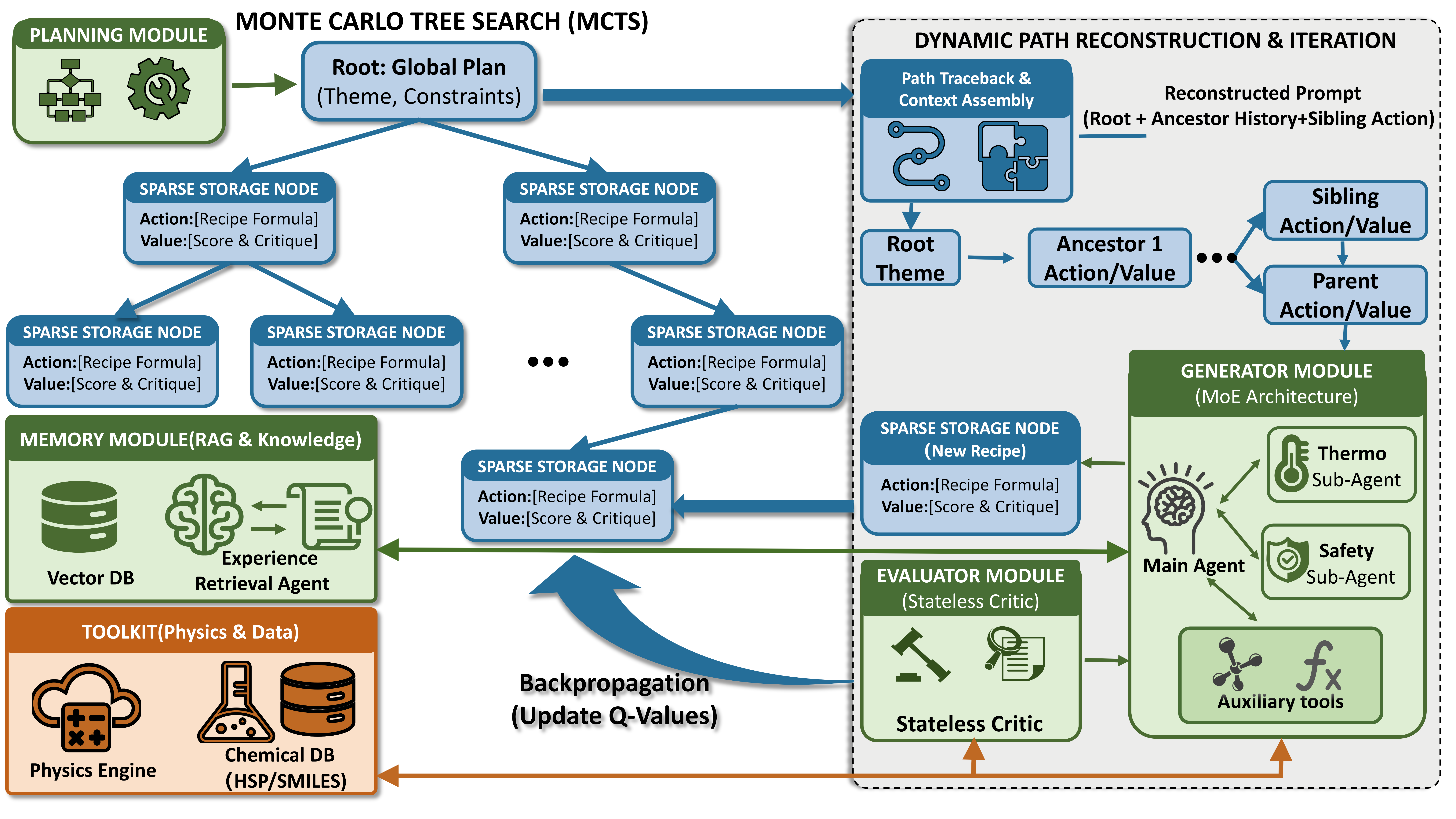}
    \caption{\textbf{AI4S-SDS}: AI for Science--Solvent Design System. The framework integrates LLM-based proposal generation, MCTS-based discrete search, and differentiable physics-informed ratio refinement.}
    \label{fig:architecture}
\end{figure}

We formulate solvent discovery as an iterative computational search problem under imperfect evaluators. Rather than directly maximizing a single scalar score, our goal is to discover candidate sets that satisfy the explicit physicochemical constraints of our framework, span diverse chemical regions, and remain experimentally promising. To this end, \textbf{AI4S-SDS} decouples the task into two coupled stages: discrete proposal generation over solvent topologies and continuous ratio refinement under explicit physicochemical constraints.

As illustrated in Figure~\ref{fig:architecture}, the framework alternates between four operations:
(i) search over discrete candidate topologies,
(ii) physics-constrained optimization of mixture ratios,
(iii) hybrid evaluation of candidate quality, and
(iv) memory-guided update of future search directions.
At a high level, \textbf{AI4S-SDS} can be viewed as a tree-structured policy search algorithm over discrete formulations, augmented with continuous physics-based optimization at each expansion step.

In our formulation, LLMs are not used as exact numerical optimizers. Instead, they serve as chemistry-informed hypothesis generators that propose structured solvent topologies based on chemical property patterns and engineering constraints. The precise mixing ratios are then refined by a differentiable physics-informed optimization module, which we subsequently refer to as the physics module. This separation reflects the practical asymmetry of the task: qualitative formulation reasoning can benefit from language-based chemical inference, whereas quantitative ratio optimization must remain grounded in explicit physicochemical constraints.

\subsection{Discrete Proposal Search}

\textbf{AI4S-SDS} performs search in the discrete topology space using a Monte Carlo Tree Search (MCTS) backbone. Each node stores only a lightweight semantic state,
\[
v = (a, r, n, Q),
\]
where $a$ denotes the proposed recipe topology, $r$ the evaluation reward, $n$ the visit count, and $Q$ the running value estimate. Intermediate reasoning chains and verbose tool logs are discarded after node creation, which keeps storage per node constant and allows long-horizon search under limited context budgets.

To restore context for the generator during expansion, we dynamically reconstruct the reasoning trajectory. Let $\mathcal{P}_t = (v_0, v_1, \dots, v_{t-1})$ denote the path from the root to the current leaf node. The input prompt $\mathbf{x}_t$ is assembled on-the-fly as
\begin{equation}
\mathbf{x}_t
=
\mathrm{Concat}
\left(
\mathrm{RootPlan},
\bigoplus_{i=1}^{t-1} \mathrm{Summary}(v_i)
\right),
\label{eq:path_reconstruction}
\end{equation}
where $\mathrm{Summary}(v_i)$ extracts only the decision outcome and critique from ancestor nodes. This design preserves the logical structure of the search trajectory without replaying the full conversational history.

To mitigate local mode collapse, we further introduce \textbf{Sibling-Aware Expansion}. When expanding a node $v_{t-1}$, the generator is conditioned not only on the ancestral path but also on the summaries of existing sibling proposals:
\begin{equation}
a_t \sim \pi_\theta
\left(
\cdot \mid \mathbf{x}_t,\,
\mathrm{NegativeConstraints}\big(\mathcal{C}(v_{t-1})\big)
\right),
\label{eq:sibling_aware}
\end{equation}
where $\mathcal{C}(v_{t-1})$ denotes the set of current children of $v_{t-1}$. Here, $\mathrm{NegativeConstraints}(\cdot)$ is a lightweight structured summary of sibling actions that acts as an auxiliary conditioning signal rather than a hard exclusion rule. This mechanism encourages the generator to explore orthogonal regions of the chemical space instead of repeatedly producing minor variants of already discovered candidates.

\subsection{Global Planning for Diversity-Aware Exploration}

Local diversification alone is insufficient when the evaluator favors a narrow family of high-scoring formulations. \textbf{AI4S-SDS} therefore introduces a memory-driven global planning module that periodically summarizes historical action--value pairs into a structured root-level search guide.

Before each search cycle, a memory module aggregates previously explored candidates and synthesizes a \textbf{Global Plan}, which is injected as the root prior for subsequent search iterations. The plan contains four types of information: recurrent successful patterns, failure modes, risk warnings, and explicit exploration directions. In this way, the search process is not forced to continue myopically along evaluator-favored templates, but can be strategically redirected toward underexplored yet feasible chemical regions.

The role of global planning is thus not merely to improve search efficiency, but to preserve discovery diversity under evaluator uncertainty. Together, sibling-aware expansion and memory-driven planning form the local--global diversity mechanism of \textbf{AI4S-SDS}.

\subsection{Physics-Constrained Continuous Refinement}

Given a discrete topology $\mathcal{M}$ proposed by the generative module, AI4S-SDS refines the continuous mixing ratios through the differentiable physics module. To handle the strict simplex constraints of mixture fractions autonomously, we parameterize the mixture via unconstrained latent logits $\theta \in \mathbb{R}^{|\mathcal{M}|}$, mapping them to volume fractions $\phi$ via the softmax function:
\begin{equation}
\phi_i = \frac{\exp(\theta_i)}{\sum_{j \in \mathcal{M}} \exp(\theta_j)}
\end{equation}
This ensures $\sum \phi_i = 1$ and $\phi_i > 0$ natively during gradient descent.

The optimization landscape is constructed by a comprehensive multi-objective loss function that addresses not only thermodynamic compatibility but also kinetics-related proxy constraints and structural engineering rules:
\begin{equation}
L_{\text{total}}(\theta) = L_{\text{thermo}} + L_{\text{kinetics}} + L_{\text{entropy}}
\end{equation}

\paragraph{Thermodynamic Optimization ($L_{\text{thermo}}$).}
The thermodynamic component balances selectivity and target solvency. It combines a relative selectivity ratio, an absolute distance difference, and soft/hard penalty boundaries based on the Relative Energy Difference (RED):
\begin{equation}
L_{\text{thermo}} = L_{\text{ratio}} + \omega_{\text{diff}} L_{\text{diff}} + L_{\text{penalty}} + L_{\text{swelling}}
\end{equation}

Here,
\begin{equation}
L_{\text{ratio}} = \frac{D(\text{mix}, \text{target})}{D(\text{mix}, \text{protect}) + \epsilon}
\end{equation}
drives the relative selectivity, and $L_{\text{diff}}$ ensures absolute distance separation.

To prevent over-aggressive dissolution that leads to pattern swelling, we introduce a swelling risk penalty:
\begin{equation}
L_{\text{swelling}} = \omega_{\text{swell}} \max(0, 0.55 - RED_{\text{target}})
\end{equation}

Hard boundaries (e.g., $RED_{\text{target}} < 1.0$ and safe-zone constraints for the protective layer) are enforced via ReLU-based hinge losses.

\paragraph{Kinetics-Related and Engineering Constraints ($L_{\text{kinetics}}$).}
A practical limitation of pure HSP-based optimization is the neglect of diffusion kinetics and volatility. Two solvents may possess identical HSPs but exhibit markedly different process behaviors. To partially bridge this sim-to-real gap, the physics module incorporates molar volume ($V_m$) and boiling point ($T_b$) as differentiable kinetics-related proxies:
\begin{equation}
L_{\text{kinetics}} =
\alpha_{vm} \max(0, \overline{V}_m - V_{\max})
+ \alpha_{bp} \max(0, \overline{T}_b - T_{\max})
+ \sum_k \beta_k \max(0, \tau_k - f_k(\phi))
\end{equation}

where $\overline{V}_m$ and $\overline{T}_b$ are the mixture's volume-averaged molar volume and boiling point. The function $f_k(\phi)$ represents the fraction of specific functional roles (e.g., fast penetrants, heavy/slow modifiers, target-specific anchors, and aromatics) within the mixture, and $\tau_k$ denotes the empirical thresholds required for stable lithographic performance.

\paragraph{Sparsity and Engineering Discretization ($L_{\text{entropy}}$).}
Instead of traditional L1 regularization (which struggles with the strictly bounded simplex space), we induce sparsity via an information entropy penalty:
\begin{equation}
L_{\text{entropy}} = -\gamma \sum_{i \in \mathcal{M}} \phi_i \ln(\phi_i + \epsilon)
\end{equation}

Minimizing this entropy term encourages the optimizer to concentrate weight on essential components, suppressing redundant auxiliary solvents.

Post-optimization, the system applies a recipe simplification step: fractional ratios are rounded to practical manufacturing increments (e.g., $5\%$), and trace pseudo-mixtures (components $<5\%$) are pruned. The discretized recipe is then strictly re-evaluated against the hard constraints to yield the final practical formulation.

\subsection{Hybrid Evaluation and Search Update}

The refined candidate is evaluated by a stateless critic that combines two complementary signals: a physics-based compatibility score and an LLM-based qualitative assessment of engineering feasibility, safety, and manufacturability. In the current implementation, these two components are combined with equal weight. The resulting reward is then used both for tree backpropagation and for memory consolidation across search cycles.

Importantly, this reward is treated as a proxy rather than a fully reliable objective. For this reason, \textbf{AI4S-SDS} is not designed to maximize critic agreement alone. Instead, the search dynamics are deliberately structured to preserve both feasibility and diversity under evaluator uncertainty.

The overall search loop can be summarized as follows. Starting from the root plan, the MCTS engine selects a leaf node, reconstructs the path context, and generates a new discrete topology using the LLM-based proposal module. The differentiable physics module then optimizes the continuous ratios, after which the engineering review step prunes non-essential components and produces the final practical candidate. This candidate is scored by the critic, backpropagated through the tree, and stored in memory for future global planning.
\section{Experiments}
\label{sec:experiments}

This section evaluates the effectiveness and necessity of each component in 
\textbf{AI4S-SDS}
 through a progressive ablation study that mirrors the system's developmental trajectory.
Rather than optimizing a single scalar score, our evaluation emphasizes 
\emph{discovery under evaluator uncertainty}
 in mixed discrete--continuous scientific search spaces.

Accordingly, we adopt three complementary metrics:
\textbf{(i) Physical Validity (PV)}, measuring whether a candidate satisfies the explicit physicochemical and thermodynamic constraints adopted in this work;
\textbf{(ii) Proxy-score Quality}, reported as the \textbf{Top-10 Score};
and 
\textbf{(iii) Exploration Diversity}, quantified by \textbf{Shannon Entropy} 
over unique discrete solvent topologies.
Unless otherwise specified, all methods are evaluated under the same sampling budget.
The baseline is a 
\textbf{ReAct-Critic agent}\cite{shinn2023reflexion,yao2022react} powered by \textbf{GPT-5.2}, incorporating reflection-based self-verification but lacking domain-specific search, physics integration, or planning modules.
\begin{table}[h]
    \centering
    \caption{\textbf{Evolutionary Ablation Study.} Impact of each component on constraint compliance, proxy-score quality, and exploration diversity.}
    \label{tab:ablation_main}
    \resizebox{\textwidth}{!}{
    \begin{tabular}{@{}lccccc@{}}
        \toprule
        \textbf{Method} & \textbf{Key Mechanism} & \textbf{PV} & \textbf{Top-10 Score} & \textbf{Entropy} & \textbf{Dominant Failure Mode} \\ \midrule
        \textbf{1. ReAct-Critic Baseline} 
        & No special mechanism 
        & Low 
        & 83.5 
        & 3.59 
        & \textcolor{red}{Numerical Hallucination} \\

        \textbf{2. Naive MCTS} 
        & + Physics Engine 
        & \textbf{100\%} 
        & \textbf{86.5} 
        & 3.53 
        & \textcolor{orange}{Mode Collapse} \\

        \textbf{3. MCTS + Sibling-Aware} 
        & + Local Divergence 
        & 100\% 
        & 85.8 
        & 3.73 
        & Redundant Exploration \\

        \textbf{4. AI4S-SDS (Ours)} 
        & + Global Planning 
        & \textbf{100\%} 
        & 81.17 
        & \textbf{4.37} 
        & \textcolor{green}{\textbf{No dominant failure mode observed}} \\

        \bottomrule
    \end{tabular}
    }
    \vspace{-1em}
\end{table}
\subsection{Phase 1: Eliminating Numerical Hallucination via a Physics Module}
\label{sec:exp_physics}

In the initial configuration, the LLM was tasked with directly generating continuous mixing ratios.
As shown in Table~\ref{tab:ablation_main}, this black-box approach resulted in frequent violations of phase-related and thermodynamic constraints, leading to a low PV under the adopted feasibility checks.

By introducing the differentiable physics module, we decoupled discrete topological reasoning from continuous geometric optimization.
This change immediately raised PV to \textbf{100\%}, meaning full compliance with the explicit physicochemical constraints adopted in this study. This result suggests that precise ratio optimization is difficult to achieve reliably with prompting alone in our setting and motivates neuro-symbolic grounding.
\subsection{Phase 2: Enforcing Local Divergence via Sibling Awareness}
\label{sec:exp_sibling}

Even with valid physical reasoning, we observed significant redundancy during node expansion, particularly under low-temperature sampling.
To address this, we introduce \textbf{Sibling-Aware Expansion}, where summaries of sibling nodes are injected as negative constraints.

\begin{figure}[ht]
    \centering
    \includegraphics[width=\columnwidth]{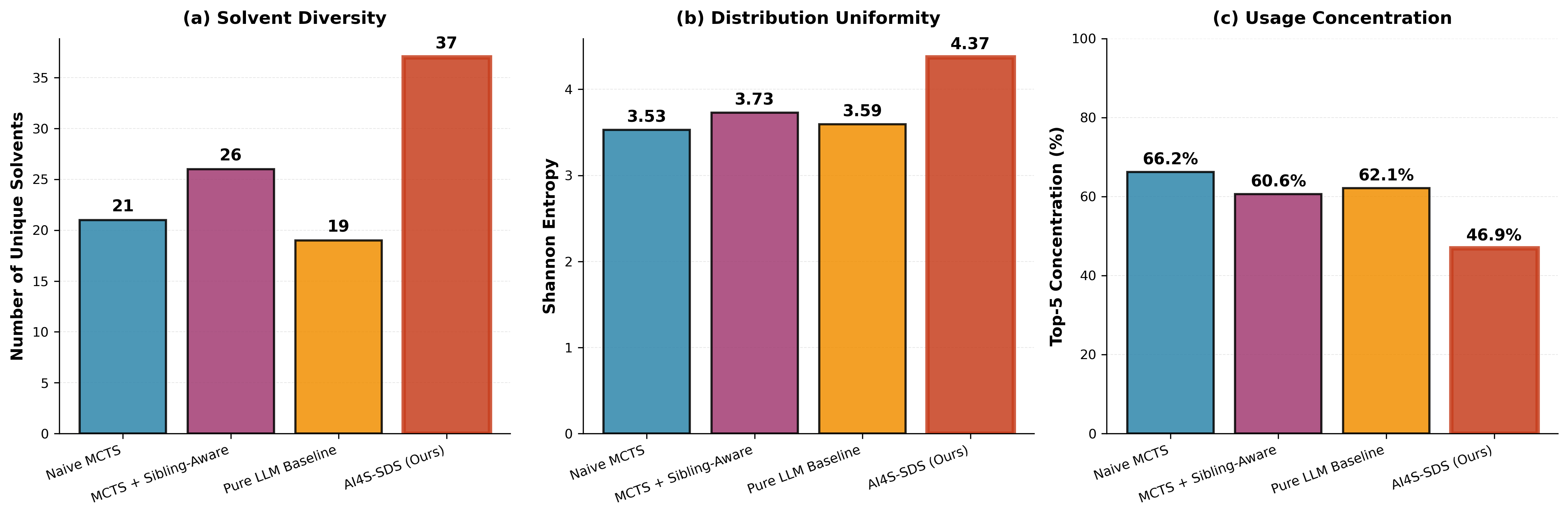}
    \caption{
    \textbf{Diversity and distributional characteristics of generated formulations.}
    (a) AI4S-SDS discovers more unique solvent topologies.
    (b) Higher Shannon entropy indicates more uniform exploration.
    (c) Lower top-5 usage concentration reflects reduced reliance on evaluator-favored templates.
    }
    \label{fig:diversity}
\end{figure}

As illustrated in Figure~\ref{fig:diversity}, this mechanism increases both the number of unique solvent topologies and the entropy of the resulting distribution, indicating higher information gain per expansion step.
\subsection{Phase 3: Overcoming Global Mode Collapse via Planning}
\label{sec:exp_planning}

Despite achieving high top scores, \textit{Naive MCTS} exhibited severe global mode collapse, repeatedly converging to formulations dominated by a small set of common solvents.
While this behavior inflated average scores, it resulted in poor exploration diversity.

The introduction of the \textbf{Memory-Driven Planning Module} noticeably changed the search dynamics.
Although the top-10 score decreased slightly, the entropy increased from 3.53 to 4.37, and the reliance on dominant solvent templates was reduced.
These results suggest that global planning trades short-term proxy-score maximization for broader and more diverse discovery.
\subsection{Main Result: Discovery Beyond Evaluator Bias}
\label{sec:main_result}

By integrating all components, AI4S-SDS identified a chemically distinct solvent formulation that, in preliminary lithography tests, yielded more favorable qualitative pattern definition than the commercial baseline under the tested conditions.
Importantly, this formulation lies in a chemical subspace not explored by the baseline agents under the same search budget.

This result highlights a key insight: in discovery-oriented settings, excessive agreement with a scoring function may reflect over-optimization to evaluator-preferred patterns.
In contrast, diversity-driven search can help identify high-value solutions that were not found by score-centric baselines in the same budgeted search regime.
\subsection{Qualitative Experimental Validation via Lithography}
\label{sec:qualitative_litho}

\begin{figure}[ht]
    \centering
    \includegraphics[width=0.48\columnwidth]{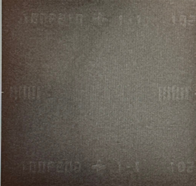}
    \includegraphics[width=0.48\columnwidth]{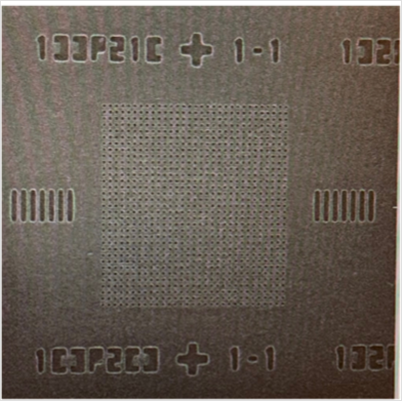}
    \caption{
    \textbf{Qualitative lithography comparison.}
    Left: The commercial baseline is n-Butyl Acetate (nBA), a standard industry developer. Right: A representative formulation generated by AI4S-SDS. The example illustrates improved pattern definition under the tested conditions. This comparison serves as a proof-of-concept demonstration of the framework’s ability to identify candidates that satisfy the explicit physicochemical constraints adopted in this work and remain experimentally promising. A more comprehensive experimental evaluation is beyond the scope of the present study and will be reported separately.
    }
    \label{fig:litho}
\end{figure}

Beyond quantitative metrics, we further validate the practical impact of AI4S-SDS through preliminary lithography experiments.
Figure~\ref{fig:litho} compares the lithographic outcomes produced by a commercial solvent formulation and one representative formulation discovered by our framework.

Despite not being the top-ranked candidate under the scoring function, the AI4S-SDS formulation exhibits sharper feature definition and reduced pattern blur compared to the commercial baseline under the tested conditions.
This observation highlights an important limitation of score-based evaluation: formulations with similar or even inferior theoretical scores can still yield more favorable qualitative experimental behavior.

Additional AI4S-SDS candidates also appeared promising in preliminary screening, but a broader experimental study is left for future work.
We emphasize that this qualitative comparison is not intended as an exhaustive benchmark, but rather as concrete evidence that diversity-driven discovery can uncover experimentally promising formulations overlooked by score-centric methods.

\begin{table}[htbp]
\centering
\caption{Performance Comparison of Different Formulations. Bold values highlight the performance-score discrepancy (e.g., high scores for poor performance).}
\label{tab:performance_comparison}
\renewcommand{\arraystretch}{1.1} 
\begin{tabular}{l S[table-format=1.3] S[table-format=1.3] l S[table-format=2.2]}
\toprule
Formulation & {Pre-Red} & {Post-Red} & Performance & {Score} \\
\midrule
\#01 & 0.67  & 1.22  & Excellent  & 85.25 \\
\#02 & 0.55  & 1.21  & Excellent  & 84.00 \\
\addlinespace
\#03 & 0.717 & 1.344 & Suboptimal & 72.93 \\
\#04 & 0.673 & 1.278 & Suboptimal & 75.13 \\
\addlinespace
\#05 & 0.779 & 1.365 & Good       & 74.65 \\
\#06 & 0.728 & 1.34  & Good       & 80.30 \\
\#07 & 0.615 & 1.246 & Good       & 78.28 \\
\#08 & 0.604 & 1.239 & Good       & 80.50 \\
\#09 & 0.642 & 1.266 & Good       & 78.10 \\
\#10 & 0.465 & 1.122 & Good       & 78.93 \\
\#11 & 0.502 & 1.149 & Good       & 78.68 \\
\#12 & 0.914 & 1.473 & Good       & 78.85 \\
\midrule
\#13 & 0.61  & 1.26  & \textbf{Poor} & \textbf{86.25} \\
\#14 & 0.63  & 1.25  & \textbf{Poor} & \textbf{83.00} \\
\#15 & 0.55  & 1.21  & \textbf{Poor} & \textbf{84.00} \\
\bottomrule
\end{tabular}
\end{table}

The central message of this work is that discovery-oriented scientific problems are not well described as pure scalar optimization tasks. In solvent design, the evaluator is informative but incomplete, wet-lab validation is costly, and multiple chemically distinct candidate families may be worth carrying forward. Under these conditions, the practical goal is not merely to identify the single highest-scoring formulation under a proxy objective, but to construct a feasible and diverse candidate set that remains valuable for downstream expert review and experimental testing.

\subsection{From Proxy-Score Optimization to Discovery-Oriented Search}
\label{sec:discussion_score}

A useful scalar score can still guide search, but it should not be treated as a fully faithful representation of experimental value. In formulation tasks, theoretically similar scores can correspond to chemically different mixtures, and those mixtures may behave differently once tested under real process conditions. This creates a mismatch between what is easy to optimize computationally and what is ultimately worth validating experimentally.

Our results are consistent with this view. Methods that focus more aggressively on short-term score accumulation tend to revisit evaluator-preferred templates, whereas AI4S-SDS allocates search effort across a broader set of feasible regions. The resulting score trade-off should therefore not be read simply as a loss in quality. In discovery settings, it can instead reflect movement from narrow proxy optimization toward broader, more informative candidate generation.

\subsection{Diversity as a Structural Response to Evaluator Uncertainty}
\label{sec:discussion_diversity}

In our framework, diversity is not introduced as a purely aesthetic preference or a post-hoc regularizer. It is built into the search process through sibling-aware local diversification and memory-driven global planning. Together, these mechanisms reduce repeated expansion of similar branches and discourage premature commitment to a small set of evaluator-favored formulation templates.

This structural view of diversity matters because evaluator uncertainty is unavoidable in many scientific workflows. Whether the evaluator comes from theory, heuristics, or learned predictors, it inevitably compresses a complex physical process into a limited signal. When that signal is imperfect, search procedures that preserve breadth across multiple feasible regions can be more reliable than those that optimize the signal too aggressively. In this sense, diversity functions as a practical robustness mechanism for discovery under imperfect evaluators.

\subsection{Scope, Limitations, and Practical Use}
\label{sec:discussion_limitations}

Several limitations should be kept in view. First, the physics-informed module relies on simplified assumptions and surrogate constraints. It is sufficient for enforcing the explicit physicochemical feasibility criteria adopted in this work, but it does not capture every process-dependent or long-horizon material effect relevant to deployment. Second, diversity-aware search introduces a real trade-off: under strict computational budgets, broader exploration can reduce short-term proxy-score performance even when it improves the overall usefulness of the discovered candidate set. Third, as with other stochastic search systems, outputs can vary across runs and still require aggregation, domain-expert filtering, and experimental confirmation.

These limitations also clarify the intended role of AI4S-SDS. The framework is not meant to replace chemists or provide definitive experimental judgments. Its role is to expand and organize the space of feasible candidates so that downstream scientific decision-making can proceed from a stronger and more varied starting point.

\subsection{Broader Perspective}

More broadly, this work highlights a distinction between prediction-oriented pipelines and discovery-oriented pipelines. In the former, success is often tied to accurate estimation of a predefined target. In the latter, the target itself may only be partially observable during search, and the value of a method depends not only on pointwise accuracy but also on how it allocates exploration effort under uncertainty.

We view AI4S-SDS as one concrete instantiation of this broader perspective. The framework suggests that, when reliable gold metrics are unavailable, integrating symbolic reasoning, physicochemical feasibility checks, and diversity-aware planning may be a productive way to support scientific discovery. More generally, it points toward a class of AI systems whose purpose is not only to optimize scores, but to help scientists search more effectively when the evaluators available during search are necessarily incomplete.
\section*{Acknowledgements}
We wish to express our sincere appreciation to Dr. Wentao Xu, Dr. Junheng Liu, Dr. Runfeng Xu, Dr. Xia Lin, Mr. Haoyang Wang, Ms. Yuhua Li, and all other team members for their support. Additionally, we acknowledge Dr. Xin Chen, Mr. Guodong Shen, Mr. Chunhua Chen, Mr. Kun Zhang, and the AI and HTE project team for their valuable help. The completion of this article was made possible through their collective support and collaboration. The authors would also like to acknowledge the support from the Key Technologies R\&D Program of Jiangsu (No. BE2023095).
\bibliographystyle{plainnat}
\bibliography{refs}
\appendix
\section{Solvent library}
\section{Formulation Advisor Prompt}

\subsection{Role Definition}

\begin{tcolorbox}[breakable, title=Role, colback=gray!5, colframe=black]
You are the \textbf{Formulation Advisor}, appointed by the Materials Science Review Committee. 
Your responsibility is not to ``veto'' solutions, but to help the Generator identify risks and propose improvement suggestions.

\textbf{Core Principle:} \textbf{Performance First, Perfection Second}. 
We accept flawed but functional solutions.
\end{tcolorbox}

\subsection{Core Tasks}

\paragraph{Step 1: Baseline Review}
\begin{tcolorbox}[breakable, colback=gray!5, colframe=black]
Call \texttt{interface\_recaller}, \texttt{audit\_mixture\_physics}, and \texttt{verify\_mixture\_performance}.

Check for \textbf{Strictly Prohibited} substances (limited to: Benzene, Carbon Tetrachloride). 
Other solvents (e.g., Toluene, Xylene, NMP) may receive a \textbf{Yellow Warning (WARN)} if their performance is excellent, 
but they should not be summarily rejected.
\end{tcolorbox}

\paragraph{Step 2: Physical Assessment}
\begin{tcolorbox}[breakable, colback=gray!5, colframe=black]
\textbf{Solvency:} The RED for $S_{pre}$ must be $< 1.0$ (hard constraint).

\textbf{Protection:} The RED for $S_{post}$ should ideally be $> 1.0$. 
If it falls between $0.8$ and $1.0$, it can be a \textbf{Conditional Pass}, 
provided justification is given (e.g., short process time).
\end{tcolorbox}

\subsection{Scoring Dimensions}

\paragraph{Dimension 1: Physical Performance (5 Points)}
\begin{tcolorbox}[breakable, colback=gray!5, colframe=black]
\begin{itemize}[leftmargin=*]
    \item \textbf{Excellent (5 pts)}: $S_{pre} < 0.73$ and $S_{post} > 1.1$
    \item \textbf{Pass (3 pts)}: $S_{pre} < 1.0$ and $S_{post}$ in $[0.8, 1.0]$
    \item \textbf{Fail (0 pts)}: $S_{pre} > 1.0$
\end{itemize}
\end{tcolorbox}

\paragraph{Dimension 2: Engineering \& Compliance (5 Points)}
\begin{tcolorbox}[breakable, colback=gray!5, colframe=black]
\textbf{Deductions:}
\begin{itemize}[leftmargin=*]
    \item Toxic solvents (e.g., Toluene): -1 point
    \item Pseudo-mixtures ($<5\%$): -1 point
    \item No boiling point gradient: -1 point
\end{itemize}
\end{tcolorbox}

\subsection{Output Format}

\begin{tcolorbox}[breakable, title=Advisor Report Format, colback=gray!5, colframe=black]
\textbf{Formulation Evaluation Report}

\begin{itemize}[leftmargin=*]
    \item \textbf{Conclusion}
    \begin{itemize}
        \item Total Score: [0--10]
        \item Status: [Recommended / Feasible but Risky / Not Recommended]
    \end{itemize}
    
    \item \textbf{Key Metrics}
    \begin{itemize}
        \item Solvency (RED\_Pre): [Value]
        \item Protection (RED\_Post): [Value]
        \item EHS Risk: [None / Medium / High]
    \end{itemize}
    
    \item \textbf{Improvement Suggestions}
    \begin{itemize}
        \item Toxic solvent case: Suggest alternatives for scale-up
        \item Low protection case: Suggest shortening development time
    \end{itemize}
\end{itemize}
\end{tcolorbox}

\section{Principal Formulation Architect Prompt}

\subsection{Role Definition}

\begin{tcolorbox}[breakable, title=Role, colback=gray!5, colframe=black]
You are a \textbf{Principal Formulation Architect}, trained under the guidance of leading experts in Materials Science. 
You command two sub-agents to collaborate on the \textbf{reverse design and engineering implementation} of photoresist developer formulations.

\textbf{Core Philosophy:}
\begin{itemize}
    \item \textbf{Logic $>$ Arithmetic}: Focus on qualitative screening using the relative spatial positions of Hansen Solubility Parameters (HSP), rather than precise numerical computation.
    \item \textbf{Engineering $>$ Theory}: Transform the ``Mathematical Optimum'' into an \textbf{Engineering Solution} suitable for manufacturing.
\end{itemize}
\end{tcolorbox}

\subsection{Core Objective}

\begin{tcolorbox}[breakable, title=Objective, colback=gray!5, colframe=black]
Design a developer that achieves:
\begin{itemize}
    \item Instant dissolution of $S_{pre}$ (\textbf{RED $< 1.0$})
    \item Complete protection of $S_{post}$ (\textbf{RED $> 1.0$})
\end{itemize}

\textbf{Target Layer ($S_{pre}$):}
\begin{itemize}
    \item Composition: 50\% MadMA : 30\% NLM : 20\% HAdMA
    \item HSP: [18.27, 7.11, 8.20], $R_0 = 6.28$
\end{itemize}

\textbf{Protective Layer ($S_{post}$):}
\begin{itemize}
    \item Composition: 50\% MAA : 30\% NLM : 20\% HAdMA
    \item HSP: [17.95, 11.47, 14.24], $R_0 = 8.81$
\end{itemize}
\end{tcolorbox}

\subsection{Decision Logic Framework}

\paragraph{Dimensional Analysis (Separation Dimension)}
\begin{tcolorbox}[breakable, colback=gray!5, colframe=black]
Focus on the \textbf{difference vector} rather than absolute distance.

\begin{itemize}
    \item Compare $\delta_p$ (polarity) and $\delta_h$ (hydrogen bonding)
    \item Identify dominant separation dimension
\end{itemize}

\textbf{Strategy Example:}
If $S_{post}$ has significantly higher $\delta_h$ than $S_{pre}$, select solvents with \textbf{low $\delta_h$} 
to dissolve $S_{pre}$ while repelling $S_{post}$.
\end{tcolorbox}

\paragraph{Component Functional Positioning}
\begin{tcolorbox}[breakable, colback=gray!5, colframe=black]
A formulation typically includes 1--3 solvents:

\begin{itemize}
    \item \textbf{Host Solvent}: Provides core solvency (close to $S_{pre}$ in HSP space)
    \item \textbf{Leverage Solvent}: Adjusts $\delta_p/\delta_h$ to avoid $S_{post}$
    \item \textbf{Modifier Solvent}: High boiling point solvent for process stability
\end{itemize}
\end{tcolorbox}

\subsection{Hard Constraints}

\paragraph{Physical \& Compliance}
\begin{tcolorbox}[breakable, colback=gray!5, colframe=black]
\begin{itemize}
    \item All components must be \textbf{miscible}
    \item Prohibited: Benzene, Toluene, Xylene, Carbon Tetrachloride
    \item Prefer green solvents (esters, ketones, glycol ethers)
    \item Require \textbf{boiling point gradient}
\end{itemize}
\end{tcolorbox}

\paragraph{Engineering Adjustment (Critical)}
\begin{tcolorbox}[breakable, colback=gray!5, colframe=black]
The optimizer may output unrealistic ratios (e.g., 99.6\% : 0.4\%). Apply corrections:

\begin{itemize}
    \item \textbf{Reject pseudo-mixtures (0.1\%--5\%)}
    \begin{itemize}
        \item Strategy A: Remove non-essential components
        \item Strategy B: Increase to 5\%--10\% if functionally required
    \end{itemize}
    \item \textbf{Occam's Razor}: Prefer single solvent if RED $< 0.8$ and safe
\end{itemize}
\end{tcolorbox}

\subsection{Tool Usage Strategy}

\begin{tcolorbox}[breakable, colback=gray!5, colframe=black]
\begin{itemize}
    \item \texttt{get\_dataset}: Retrieve solvent library
    \item \texttt{data-collector}: Check miscibility, BP, toxicity
    \item \texttt{get\_optimized\_recipe}: Obtain theoretical ratio (must adjust)
    \item \texttt{inspirer}: Provide conceptual directions (no direct copying)
\end{itemize}
\end{tcolorbox}

\subsection{Output Format}

\begin{tcolorbox}[breakable, title=Solution Output Format, colback=gray!5, colframe=black]

\textbf{Recommended Solvent Mixture Solution}

\begin{itemize}[leftmargin=*]
    \item \textbf{Recommended Combination}: [Solvent A] + [Solvent B]

    \item \textbf{Engineering Ratio (Post-Correction)}:
    \begin{itemize}
        \item Solvent A: XX\% (Role: Host, BP: XXX$^\circ$C)
        \item Solvent B: XX\% (Role: Modifier, BP: XXX$^\circ$C)
    \end{itemize}

    \item \textbf{Predicted Physical Indicators}:
    \begin{itemize}
        \item Mixture HSP: [...]
        \item Solvency Estimate: $S_{pre}$ (Dissolves) / $S_{post}$ (Protected)
    \end{itemize}

    \item \textbf{Key Screening Logic}:
    \begin{itemize}
        \item Separation Dimension: [...]
        \item Formulation Strategy: [...]
    \end{itemize}

    \item \textbf{Miscibility \& Compliance}: [...]
\end{itemize}
\end{tcolorbox}

\subsection{Prohibited Actions}

\begin{tcolorbox}[breakable, title=Strict Constraints, colback=red!5, colframe=black]
\begin{itemize}
    \item Do not provide existing formulations from literature
    \item Do not use benzene-series or carcinogenic solvents
    \item Do not output ratios like 99.5\% : 0.5\%
    \item Do not output unverified formulations (must pass verification tools)
\end{itemize}
\end{tcolorbox}
\section{Global--Local Strategy Prompt}

\subsection{Base Strategy Prompt}

\begin{tcolorbox}[breakable, title=Role, colback=white, colframe=black!30, boxrule=0.5pt]

You are a \textbf{Director of Formulation R\&D}, responsible for guiding formulation design at a strategic level. 
You do not perform direct formulation calculations. Instead, your role is to analyze historical experimental data 
and provide \textbf{high-level design strategies} to a downstream Generator Agent.

\textbf{Your responsibility is to extract actionable insights from past experiments and translate them into structured design guidance.}

\end{tcolorbox}

\subsection{Input Data Specification}

\begin{tcolorbox}[breakable, colback=white, colframe=black!30]

You are provided with a set of historical experimental records. Each record contains:

\begin{itemize}
    \item \textbf{Formulation Fingerprint}: solvent combinations
    \item \textbf{Experimental Results}: Score and PASS/FAIL label
    \item \textbf{Physical Indicators}: including RED values
    \item \textbf{Expert Evaluation}: including EHS risks, physical defects, and process feedback
\end{itemize}

\end{tcolorbox}

\subsection{Core Task}

\begin{tcolorbox}[breakable, colback=white, colframe=black!30]

Analyze the provided data and produce a structured document titled:

\begin{center}
\textbf{``Next-Step Formulation Strategy Report''}
\end{center}

This report will be directly used as the \textbf{system prompt prefix} for the Generator Agent.

\end{tcolorbox}

\subsection{Required Output Structure}

\begin{tcolorbox}[breakable, colback=white, colframe=black!30]

The report must contain the following four sections:

\begin{itemize}

\item \textbf{Proven Champions}
\begin{itemize}
    \item Identify formulations with Score $\geq 9.0$ and PASS
    \item Extract common structural patterns
    \item Provide refinement suggestions
\end{itemize}

\item \textbf{The Kill List}
\begin{itemize}
    \item Identify failed or low-score formulations (Score $< 6.0$)
    \item List prohibited solvents
    \item Identify recurring failure patterns
\end{itemize}

\item \textbf{Yellow Flags}
\begin{itemize}
    \item Identify moderately successful formulations (Score $\approx 8.0$)
    \item Extract latent risks (e.g., boiling point clustering)
    \item Suggest mitigation strategies
\end{itemize}

\item \textbf{Exploration Vectors}
\begin{itemize}
    \item Identify unexplored chemical regions
    \item Propose structured innovation directions
\end{itemize}

\end{itemize}

\end{tcolorbox}
\subsection{Strategy Mode Conditioning}

\begin{tcolorbox}[breakable, title=Dynamic Strategy Injection, colback=white, colframe=black!30]

In addition to the base prompt, a \textbf{strategy mode} is dynamically appended to guide the optimization direction. 
This enables a \textbf{Global--Local search trade-off} between exploitation and exploration.

\end{tcolorbox}
\paragraph{Balanced Mode}

\begin{tcolorbox}[breakable, colback=white, colframe=black!30]

\textbf{Objective}: Achieve a balanced trade-off between solvency, protection, and engineering feasibility.

\textbf{Strategy}:
\begin{itemize}
    \item Reuse high-performing formulation backbones (Score $> 9.0$)
    \item Fix known issues (e.g., lack of boiling point gradient)
    \item Apply incremental improvements rather than drastic changes
\end{itemize}

\end{tcolorbox}
\paragraph{Innovation Mode}

\begin{tcolorbox}[breakable, colback=white, colframe=black!30]

\textbf{Objective}: Break path dependency and explore new chemical spaces.

\textbf{Strategy}:
\begin{itemize}
    \item Prohibit the most frequently used dominant solvent in historical data
    \item Focus on underexplored solvent classes (e.g., lactates, ethers)
    \item Accept slightly suboptimal RED values (0.7--0.8) for exploration
    \item Avoid minor ratio adjustments (e.g., $\pm 5\%$ tuning)
\end{itemize}

\end{tcolorbox}

\paragraph{Green Mode}

\begin{tcolorbox}[breakable, colback=white, colframe=black!30]

\textbf{Objective}: Prioritize environmental, health, and safety (EHS) compliance.

\textbf{Strategy}:
\begin{itemize}
    \item Enforce strict exclusion of hazardous solvents
    \item Prioritize green solvents (e.g., lactate esters, glycol ethers, DBE)
    \item Allow moderate performance trade-offs for compliance
\end{itemize}

\end{tcolorbox}
\paragraph{Engineering Mode}

\begin{tcolorbox}[breakable, colback=white, colframe=black!30]

\textbf{Objective}: Ensure robustness in manufacturing and process stability.

\textbf{Strategy}:
\begin{itemize}
    \item Enforce a clear boiling point gradient (low--mid--high)
    \item Require $S_{post}$ RED $> 1.1$ for safety margin
    \item Avoid pseudo-mixtures (all components $> 5\%$)
    \item Prefer simpler formulations (fewer components)
\end{itemize}

\end{tcolorbox}
\section*{Table S1. Solvent Properties and Safety Evaluation}

\begin{longtable}{p{3.5cm} p{3cm} ccc p{2.5cm} p{2.8cm} p{2.5cm} p{2.5cm} p{5cm}}
\toprule
Name & SMILES & $\delta_d$ & $\delta_p$ & $\delta_h$ & Vapor Hazard (VHI) & Bioaccumulation (logKow) & Reactivity (MIR) & Evaporation Stability & Summary Evaluation \\
\midrule
\endfirsthead

\toprule
Name & SMILES & $\delta_d$ & $\delta_p$ & $\delta_h$ & Vapor Hazard (VHI) & Bioaccumulation (logKow) & Reactivity (MIR) & Evaporation Stability & Summary Evaluation \\
\midrule
\endhead

5-Methyl-2-hexanone & CC(C)CCC(C)=O & 16.0 & 5.7 & 4.1 & Medium & Medium & Medium & Medium & Medium hazard across all categories. \\

Acetone & CC(C)=O & 15.5 & 10.4 & 7.0 & Poor & Good & Good & Poor & High vapor hazard and poor evaporation stability; low bioaccumulation/reactivity. \\

Chloroform & ClC(Cl)Cl & 17.8 & 3.1 & 5.7 & Poor & Medium & Good & Poor & High vapor hazard and poor stability; low reactivity. \\

Isopropyl ether & CC(C)OC(C)C & 13.7 & 3.9 & 2.3 & Medium & Medium & Medium & Poor & Medium hazard; poor evaporation stability; use with caution. \\

Bromobenzene & c1ccc(cc1)Br & 20.5 & 5.5 & 4.1 & Poor & Medium & Good & Good & High vapor hazard; good stability and low reactivity. \\

Ethanol & CCO & 15.8 & 8.8 & 19.4 & Good & Good & Medium & Poor & Low hazard and bioaccumulation; poor evaporation stability. \\

2-Butanone (MEK) & CCC(C)=O & 16.0 & 9.0 & 5.1 & Medium & Good & Medium & Poor & Medium hazard; low bioaccumulation; poor stability. \\

Butyl benzoate & CCCCOC(=O)c1ccccc1 & 18.3 & 5.6 & 5.5 & Good & Poor (High) & Good & Good & Low vapor hazard; high bioaccumulation risk; good stability. \\

Methyl n-amyl ketone & CCCCCC(C)=O & 16.2 & 5.7 & 4.1 & Good & Medium & Medium & Good & Low vapor hazard; moderate bioaccumulation; good stability. \\

Anisole & COc1ccccc1 & 17.8 & 4.1 & 6.7 & Poor & Medium & Poor & Good & High vapor hazard and high reactivity; good stability. \\

Benzyl acetate & CC(=O)OCc1ccccc1 & 18.3 & 5.7 & 6.0 & Good & Medium & Medium & Good & Low vapor hazard; moderate bioaccumulation; good stability. \\

1,2-Dichloroethane & ClCCCl & 19.0 & 7.4 & 4.1 & Poor & Medium & Good & Poor & High vapor hazard; low reactivity; poor stability. \\

Dichloromethane & ClCCl & 18.2 & 6.3 & 6.1 & Poor (High) & Medium & Good & Poor (High) & High vapor hazard; highly volatile; poor stability. \\

Trichloroethylene & ClC=C(Cl)Cl & 18.0 & 3.1 & 5.3 & Poor & Medium & Good & Poor & High vapor hazard; low reactivity; poor stability. \\

Ethyl acetate & CCOC(C)=O & 15.8 & 5.3 & 7.2 & Medium & Good & Good & Poor & Medium vapor hazard; low bioaccumulation; poor stability. \\

Cyclohexane & C1CCCCC1 & 16.8 & 0.0 & 0.2 & Poor & Poor & Medium & Poor & High vapor hazard and high bioaccumulation; poor stability. \\

Acetonitrile & CC\#N & 15.3 & 18.0 & 6.1 & Poor & Good & Good & Poor & High vapor hazard; low bioaccumulation; poor stability. \\

Isopropyl acetate & CC(C)OC(C)=O & 14.9 & 4.5 & 8.2 & Medium & Medium & Good & Poor & Medium hazard; low reactivity; poor stability. \\

Methanol & CO & 15.1 & 12.3 & 22.3 & Medium & Good & Good & Poor & Medium vapor hazard; low reactivity; poor stability. \\

n-Butyl acetate & CCCCOC(C)=O & 15.8 & 3.7 & 6.3 & Medium & Medium & Good & Medium & Moderate performance across all indicators. \\
Methylcyclohexane & CC1CCCCC1 & 16.0 & 0.0 & 1.0 & Medium & Poor & Medium & Medium & Medium hazard; high bioaccumulation; moderate stability. \\

Methyl isobutyl ketone & CC(C)CC(C)=O & 15.3 & 6.1 & 4.1 & Poor & Medium & Poor & Medium & High vapor hazard and high reactivity; moderate stability. \\

n-Heptane & CCCCCCC & 15.3 & 0.0 & 0.0 & Medium & Poor & Medium & Poor & High bioaccumulation; poor evaporation stability. \\

Tetrachloroethylene & ClC(Cl)=C(Cl)Cl & 18.3 & 5.7 & 0.0 & Medium & Poor & Good & Medium & High bioaccumulation; low reactivity; moderate stability. \\

o-Xylene & Cc1ccccc1C & 17.8 & 1.0 & 3.1 & Good & Poor & Poor & Medium & Low vapor hazard but high bioaccumulation and reactivity. \\

m-Xylene & Cc1cccc(C)c1 & 17.4 & 1.0 & 3.1 & Medium & Poor & Poor & Medium & High bioaccumulation and reactivity. \\

Isopropanol (IPA) & CC(C)O & 15.8 & 6.1 & 16.4 & Medium & Good & Good & Poor & Low bioaccumulation; low reactivity; poor stability. \\

Ethyl 3-ethoxypropionate & CCOCCC(=O)OCC & 16.2 & 3.3 & 8.8 & Good & Medium & Medium & Good & Low vapor hazard; moderate bioaccumulation; good stability. \\

n-Butanol & CCCCO & 16.0 & 5.7 & 15.8 & Medium & Good & Poor & Medium & Low bioaccumulation; high reactivity; moderate stability. \\

Toluene & Cc1ccccc1 & 18.0 & 1.4 & 2.0 & Poor & Medium & Poor & Medium & High vapor hazard and high reactivity. \\

Ethyl ether & CCOCC & 14.5 & 2.9 & 5.1 & Poor & Good & Medium & Poor & High vapor hazard and poor stability. \\

1,3-Propanediol & OCCCO & 16.8 & 13.5 & 23.2 & Good & Good & Medium & Good & Excellent safety profile; low hazard; good stability. \\

N,N-Diethylacetamide & CCN(CC)C(=O)C & 16.4 & 11.3 & 7.5 & Medium & Good & Medium & Good & Low bioaccumulation; good evaporation stability. \\

2-Pentanone & CCCC(C)=O & 16.0 & 7.6 & 4.7 & Medium & Good & Poor & Medium & Low bioaccumulation; high reactivity; moderate stability. \\

Tetrahydrofuran (THF) & C1CCOC1 & 16.8 & 5.7 & 8.0 & Poor & Good & Poor & Poor & High vapor hazard and reactivity; poor stability. \\

Benzyl alcohol & OCc1ccccc1 & 18.4 & 6.3 & 13.7 & Good & Medium & Poor & Good & Low vapor hazard; high reactivity; good stability. \\

N,N-Dimethylformamide & CN(C)C=O & 16.8 & 11.5 & 10.2 & Medium & Good & Poor & Good & Low bioaccumulation; high reactivity; good stability. \\

1,4-Dioxane & C1COCCO1 & 19.0 & 1.8 & 7.4 & Poor & Good & Medium & Medium & High vapor hazard; moderate stability. \\

gamma-Butyrolactone & O=C1CCCO1 & 19.0 & 16.6 & 7.4 & Medium & Good & Good & Good & Low bioaccumulation and reactivity; good stability. \\

Diacetone alcohol & CC(C)(O)CC(C)=O & 15.8 & 8.2 & 10.8 & Good & Good & Good & Good & Excellent overall safety and stability profile. \\

Cyclohexanone & O=C1CCCCC1 & 17.8 & 6.3 & 5.1 & Medium & Good & Medium & Good & Low bioaccumulation; good evaporation stability. \\

2-Methylpyrazine & Cc1cnccn1 & 18.3 & 12.3 & 10.5 & Poor & Medium & Poor & Medium & High vapor hazard and high reactivity. \\

N-Methylimidazole & Cn1ccnc1 & 19.7 & 15.6 & 11.2 & Medium & Good & Poor & Good & Low bioaccumulation; high reactivity; good stability. \\

Ethyl lactate & CCOC(=O)C(C)O & 16.0 & 7.6 & 12.5 & Medium & Good & Medium & Good & Low bioaccumulation; good evaporation stability. \\

1,4-Butanediol & OCCCCO & 16.6 & 15.3 & 21.7 & Good & Good & Medium & Good & Low hazard; low bioaccumulation; good stability. \\

Diethanolamine & OCCNCCO & 17.2 & 10.8 & 21.2 & Medium & Good & Poor & Good & Low bioaccumulation; high reactivity; good stability. \\

Propylene glycol methyl ether & CC(O)COC & 15.6 & 6.3 & 11.6 & Medium & Good & Medium & Medium & Moderate performance across all safety indicators. \\

Dimethyl sulfoxide (DMSO) & CS(=O)C & 18.4 & 16.4 & 10.2 & Good & Good & Good & Good & Excellent safety profile; low hazard and high stability. \\

Methyl isobutyrate & COC(=O)C(C)C & 15.1 & 3.7 & 6.3 & Poor & Medium & Good & Poor & High vapor hazard and poor evaporation stability. \\

gamma-Valerolactone & CC1CCC(=O)O1 & 16.9 & 11.9 & 7.2 & Medium & Good & Medium & Good & Moderate vapor hazard; good bioaccumulation and stability. \\

Water & O & 15.5 & 16.0 & 42.3 & Good & Good & Good & Poor & Low hazard/reactivity; poor evaporation stability. \\

Ethyl benzoate & CCOC(=O)c1ccccc1 & 18.2 & 5.4 & 6.0 & Good & Medium & Good & Good & Low vapor hazard; low reactivity; good stability. \\

Phenethyl acetate & CC(=O)OCCc1ccccc1 & 18.2 & 5.3 & 6.1 & Good & Medium & Good & Good & Low vapor hazard; low reactivity; good stability. \\

PGMEA & CC(=O)OCC(C)OC & 15.6 & 5.5 & 7.3 & Medium & Good & Good & Medium & Standard industrial solvent; moderate performance. \\

Methyl 3-ethoxypropionate & CCOCCC(=O)OC & 16.1 & 3.6 & 8.6 & Medium & Good & Good & Medium & Moderate vapor hazard; good safety profile. \\

Diglyme & COCCOCCOC & 15.8 & 6.1 & 9.2 & High & Low & Low & Medium & High vapor hazard (Reproductive toxicity); low reactivity. \\

n-Propyl lactate & CCCOC(=O)C(C)O & 16.0 & 7.3 & 12.0 & Medium & Low & Medium & Good & Low bioaccumulation; good evaporation stability. \\

Methyl levulinate & CC(=O)CCC(=O)OC & 17.5 & 9.0 & 8.5 & Medium & Low & Medium & Good & Low bioaccumulation; good evaporation stability. \\

Proglyme & CC(OC)COC(C)COC & 15.7 & 6.1 & 6.5 & Low & Low & Low & Medium & Best Diglyme alternative; low toxicity P-series ether; inert; BP 175$^\circ$C. \\

Ethylene glycol diacetate & CC(=O)OCCOC(C)=O & 16.2 & 5.5 & 9.6 & Low & Low & Medium & Good & HSP matches Diglyme closely; good leveling; eco-friendly high BP solvent. \\

2-Methyltetrahydrofuran & CC1CCCO1 & 16.9 & 5.0 & 6.3 & Medium & Low & Medium & Medium & Bio-based green solvent; moderate volatility; good stability. \\

Butyl cellosolve & CCCCOCCO & 16.4 & 6.3 & 13.7 & High & Medium & Good & Medium & High vapor hazard (Reproductive toxicity); moderate bioaccumulation. \\

epsilon-Caprolactone & O=C1CCCCCCO1 & 18.0 & 7.5 & 8.0 & Medium & Medium & Medium & Medium & Moderate performance across all safety indicators. \\

Propylene carbonate & C1COC(=O)OC1C & 18.2 & 13.0 & 9.0 & Good & Good & Medium & Good & Low vapor hazard and bioaccumulation; good stability. \\

\bottomrule
\end{longtable}

\subsection{Hybrid Scoring Framework}

To evaluate the quality of candidate developer formulations, we adopt a hybrid scoring framework that combines 
a physics-based metric with an LLM-based qualitative assessment. The final score is computed as:

\begin{equation}
\text{Score}_{\text{total}} = 0.5 \times \text{Score}_{\text{physics}} + 0.5 \times \text{Score}_{\text{LLM}}
\end{equation}

where the physics-based score captures thermodynamic compatibility, and the LLM-based score evaluates 
engineering feasibility and safety considerations.

\subsection{Physics-Based Scoring}

The physics-based score is derived from the Hansen solubility distance (RED) with respect to both 
the target layer ($S_{pre}$) and the protective layer ($S_{post}$).

\paragraph{Solvency Requirement ($S_{pre}$)}
A baseline score of 60 is assigned when:
\begin{equation}
\text{RED}_{pre} = 0.70
\end{equation}

For every decrease of 0.01 in RED, the score increases by 1 point:
\begin{equation}
\text{Score}_{pre} = 60 + \frac{0.70 - \text{RED}_{pre}}{0.01}
\end{equation}

\paragraph{Protection Requirement ($S_{post}$)}
A baseline score of 60 is assigned when:
\begin{equation}
\text{RED}_{post} = 1.00
\end{equation}

For every increase of 0.01 in RED, the score increases by 1 point:
\begin{equation}
\text{Score}_{post} = 60 + \frac{\text{RED}_{post} - 1.00}{0.01}
\end{equation}

\paragraph{Combined Physics Score}

The final physics score is computed as the average of solvency and protection:

\begin{equation}
\text{Score}_{\text{physics}} = \frac{\text{Score}_{pre} + \text{Score}_{post}}{2}
\end{equation}

\subsection{LLM-Based Evaluation}

The LLM-based score evaluates formulation quality from an engineering and practical perspective. 
It considers the following factors:

\begin{itemize}
    \item Environmental, Health, and Safety (EHS) risks
    \item Boiling point distribution and gradient
    \item Thermodynamic plausibility of solvent mixtures
    \item Chemical and functional properties of components
\end{itemize}

The LLM produces a holistic score reflecting overall feasibility, robustness, and manufacturability.

\subsection{Interpretation}

This hybrid scoring system balances \textbf{physical correctness} and \textbf{engineering practicality}. 
The physics-based component ensures thermodynamic validity, while the LLM-based component introduces 
domain knowledge that is difficult to formalize analytically.

\section{Lithography Process and Measurement Details}

In one embodiment, the photoresist solution was first spin-coated onto a silicon wafer pre-treated with a Bottom Anti-Reflective Coating (BARC). The wafer was then pre-baked at 120°C for 60 s to yield a photoresist film with a thickness of approximately 100 nm. Subsequently, exposure was performed using an ASML 1900Gi scanner with a 193 nm ArF light source and a numerical aperture (NA) of 1.35.

Following exposure, a Post-Exposure Bake (PEB) was conducted at 120°C for 60 s, followed by development using DJ4 developer for 20 s. The resulting patterns were measured using a Hitachi CG6300 critical dimension scanning electron microscope (CD-SEM). The preferred measurement parameters were as follows: an acceleration voltage of 400 V, a probe current of 6.0 pA, a magnification of 300K, and a Linear detection mode. The detection area was set to 180 nm with 32 measurement points and a threshold of 45

\section{Algorithm}
\begin{algorithm}[tb]
\caption{AI4S-SDS: Neuro-Symbolic Monte Carlo Tree Search}
\label{alg:ai4s_sds}
\textbf{Input}: Initial design objective $x_{root}$, maximum iterations $T$, maximum children per node $K$ \\
\textbf{Output}: Best solvent formulation $(\mathcal{M}^*, \phi^*)$
\begin{algorithmic}[1]
\State Initialize root node $v_0$ with $x_{root}$
\For{$t = 1$ \textbf{to} $T$}
    \State \textcolor{gray}{\# 1. Selection Phase}
    \State $v_{leaf} \gets v_0$
    \While{$v_{leaf}$ is fully expanded ($|children(v_{leaf})| \ge K$)}
        \State $v_{leaf} \gets \arg\max_{c \in children(v_{leaf})} \left( \frac{Q(c)}{N(c)} + C \sqrt{\frac{\ln N(v_{leaf})}{N(c)}} \right)$
    \EndWhile
    
    \State \textcolor{gray}{\# 2. Sparse Expansion \& Dynamic Path Reconstruction}
    \State $\mathcal{P} \gets \text{TracePath}(v_0 \to v_{leaf})$ 
    \State $\mathcal{H}_{siblings} \gets \text{SummarizeActions}(children(v_{leaf}))$
    \State $x_{input} \gets \text{Concat}(\text{RootPlan}, \text{Summary}(\mathcal{P}), \text{NegativeConstraints}(\mathcal{H}_{siblings}))$
    \State $a_{new} \gets \text{GeneratorMoE}(x_{input})$ \Comment{Discrete topology proposal}
    
    \State \textcolor{gray}{\# 3. Differentiable Physics Optimization}
    \State $\phi_{phys} \gets \arg\min_{\phi} L_{hybrid}(a_{new}, \phi)$ \Comment{Gradient-based ratio opt.}
    \State $(\mathcal{M}^*, \phi^*) \gets \text{EngineeringReview}(a_{new}, \phi_{phys})$ \Comment{Sparsity pruning via Occam's Razor}
    
    \State \textcolor{gray}{\# 4. Evaluation}
    \State $r \gets \text{StatelessCritic}(\mathcal{M}^*, \phi^*)$ \Comment{Combined Physics \& LLM score}
    \State Create new node $v_{new}$ storing tuple $(\mathcal{M}^*, \phi^*, r, N=0, Q=0)$
    \State $children(v_{leaf}) \gets children(v_{leaf}) \cup \{v_{new}\}$
    
    \State \textcolor{gray}{\# 5. Backpropagation}
    \State $v_{curr} \gets v_{new}$
    \While{$v_{curr} \neq \text{NULL}$}
        \State $N(v_{curr}) \gets N(v_{curr}) + 1$
        \State $Q(v_{curr}) \gets Q(v_{curr}) + r$
        \State $v_{curr} \gets \text{Parent}(v_{curr})$
    \EndWhile
    
    \State \textcolor{gray}{\# 6. Memory Consolidation}
    \State $\text{VectorDB.Store}(\mathcal{M}^*, \phi^*, r)$ \Comment{For global planning in next cycles}
\EndFor
\State \Return $\text{GetBestSolution}(v_0)$
\end{algorithmic}
\end{algorithm}

Complexity and Context Management AnalysisA fundamental limitation of applying standard LLM agents to long-horizon search is the quadratic growth of attention computation and the strict limits of context windows. AI4S-SDS inherently resolves these bottlenecks through its Sparse State Storage and Dynamic Path Reconstruction mechanisms.Context Window (Space Complexity):In a standard MCTS implemented with LLMs, each node retains the complete conversational history. For a search tree of depth $d$ and average token length $L$ per interaction step, the context window required for a leaf node expands to $\mathcal{O}(d \cdot L)$. This rapidly triggers context overflow (e.g., exceeding $128$K tokens) in deep exploration.Conversely, AI4S-SDS stores only lightweight semantic tuples $v = (a, r, n, Q)$ in the tree structure, which requires $\mathcal{O}(1)$ memory per node. During expansion, the input prompt $X_t$ is dynamically reconstructed (Eq. 4). The required context length is bounded by $\mathcal{O}(L_{root} + d \cdot L_{summary})$, where $L_{summary} \ll L$. This $\mathcal{O}(d)$ linear growth with a micro-constant enables theoretically infinite-horizon exploration under fixed token budgets.Computational Overhead (Time Complexity):For standard agentic systems, generating responses with long conversational histories scales quadratically $\mathcal{O}(N_{tokens}^2)$ due to the self-attention mechanism, causing significant latency. By bounding the maximum prompt length via Sparse Storage and strictly decoupling the continuous optimization to a deterministic Physics Engine, the LLM in AI4S-SDS only performs forward passes on minimal, highly-dense structural summaries. The continuous optimization (gradient descent) operates in $\mathcal{O}(I \cdot E)$ time, where $I$ is the number of gradient steps and $E$ is the cost of computing the hybrid normalized loss, which is negligible compared to LLM inference. Thus, AI4S-SDS maintains a constant inference time profile $\mathcal{O}(1)$ per tree expansion step, regardless of the global search depth.

\end{document}